%% file: acl_latex.tex
\documentclass[11pt]{article}

\usepackage[preprint]{acl}

\usepackage{times}
\usepackage{latexsym}

\usepackage[T1]{fontenc}

\usepackage[utf8]{inputenc}

\usepackage{microtype}

\usepackage{inconsolata}

\usepackage{graphicx}
\input{math_commands.tex}

\usepackage{graphicx}
\usepackage{hyperref}
\usepackage{multirow}
\usepackage{booktabs} 
\usepackage{subcaption}
\usepackage{url}
\usepackage{longtable}   
\usepackage{booktabs}    
\usepackage{array}       
\usepackage{geometry}    
\usepackage{makecell}
\title{
Exploring Multi-Temperature Strategies for Token- and Rollout-Level Control in RLVR}

\author{
 \textbf{Haomin Zhuang\textsuperscript{1}},
 \textbf{Yujun Zhou\textsuperscript{1}},
 \textbf{Taicheng Guo\textsuperscript{1}},
\\
 \textbf{Yue Huang\textsuperscript{1}},
 \textbf{Fangxu Liu\textsuperscript{2}},
 \textbf{Kai Song\textsuperscript{2}},
 \textbf{Xiangliang Zhang\textsuperscript{1}},
 \\
 \textsuperscript{1}University of Notre Dame, \textsuperscript{2}ByteDance
 \\
 \{hzhuang2, xzhang33\}@nd.edu
}

\begin{document}

\maketitle

\begin{abstract}

Reinforcement Learning has demonstrated substantial improvements in the reasoning abilities of Large Language Models (LLMs), exhibiting significant applicability across various domains. Recent research has identified that tokens within LLMs play distinct roles during reasoning tasks, categorizing them into high-entropy reasoning tokens and low-entropy knowledge tokens. Prior approaches have typically focused on restricting updates to indirectly encourage exploration, yet they do not explicitly facilitate exploratory behavior during the token generation stage itself. In this work, we introduce a complementary approach that explicitly promotes exploration during sampling by applying distinct temperature settings for different token types. Specifically, our method employs higher temperatures for reasoning tokens to actively encourage exploration, while retaining lower temperatures for knowledge tokens to maintain factual correctness. Furthermore, we systematically investigate various multi-temperature scheduling strategies and their impacts within reinforcement learning contexts. Empirical evaluations on several reasoning benchmarks demonstrate that our approach significantly enhances the reasoning performance of LLMs. The code is available at \href{https://github.com/zhmzm/Multi_Temperature_Verl.git}{\texttt{github.com/zhmzm/Multi\_Temperature\_Verl}}.

\end{abstract}

\section{Introduction}
Large Language Models (LLMs) have exhibited remarkable performance across diverse tasks, significantly enhancing capabilities in fields ranging from natural language understanding to code generation and mathematical reasoning~\citep{comanici2025gemini,bai2023qwen,dubey2024llama}. Although supervised pre-training equips LLMs with extensive world knowledge, recent advances underscore the necessity of additional post-training strategies to optimize their reasoning skills. Reinforcement Learning (RL), in particular, has emerged as a powerful technique, showing promise in effectively refining higher-order reasoning behaviors, such as logical inference and problem-solving, without fundamentally altering the foundational knowledge encoded within the models~\citep{guo2025deepseek,xu2025towards}.

Temperature scaling is a critical yet under-explored mechanism in Reinforcement Learning with Verifiable Rewards (RLVR). It directly shapes the exploration--exploitation trade-off during generation, but most prior work either overlooks temperature or applies a single, uniform value across all tokens and contexts. Such homogeneous treatment ignores the heterogeneous exploratory needs of different token types and generation stages, potentially suppressing output diversity and degrading overall quality. This motivates a more nuanced, token-level and rollout-level multiple temperature strategy to better leverage temperature scaling for enhanced RLVR.

Recent studies~\citep{Polaris2025,liu2025scaling} have acknowledged the importance of temperature in RLVR: lower temperatures favor exploitation of existing knowledge and can stabilize training, whereas higher temperatures encourage diverse sampling and exploration—up to a point, beyond which excessively high temperatures hinder the ability to sample correct answers.
Importantly, temperature during training serves a fundamentally different purpose from temperature at inference. In training, higher temperatures encourage exploration, exposing the policy to diverse reasoning paths—even when some lead to incorrect answers—which provides useful learning signals and improves robustness. In contrast, inference-time temperature merely reshapes token probabilities without influencing the underlying learning dynamics.
Selecting an appropriate temperature for the training stage, however, remains challenging. Empirical investigations  \citep{Polaris2025,he2025skywork,liu2025scaling} show that different temperatures can lead to markedly different outcomes, and early-stage trials to identify a ``best'' temperature are not necessarily robust: a temperature that appears advantageous at the beginning of training may later cause collapse, initially indistinguishable temperatures can diverge substantially in the later phase, and broad temperature sweeps are computationally expensive. In addition, prior research~\citep{cui2025entropy,wang2025beyond,wang2025stabilizing} highlights that tokens within LLM outputs possess distinct characteristics, notably distinguishing between high-entropy reasoning tokens that handle logical connections and exploratory reasoning, and low-entropy knowledge tokens that encapsulate factual and domain-specific information.  While \citet{Polaris2025} propose gradually increasing the temperature during the training stage, the choice of temperature values and the strategies for scheduling them remain unclear. More broadly, the question of how to exploit \emph{multiple} temperatures along additional dimensions of training is still largely unexplored. In this work, we take a first step by studying \emph{multi-temperature sampling per prompt}: when generating multiple rollouts for a single query, we sample with a set of temperatures rather than a fixed one. This simple hedge aims to deliver competitive performance without extra computation cost.

Building on token-level entropy observation, existing approaches exploit the distinction by either constraining updates on high-entropy reasoning tokens to stabilize training~\citep{cui2025entropy,wang2025stabilizing}, or selectively training on these tokens to indirectly encourage exploration~\citep{wang2025beyond}. However, these passive constraints have inherent limitations, particularly when dealing with low-entropy scenarios such as models already fine-tuned via supervised fine-tuning (SFT), where limited exploration can significantly reduce learning effectiveness.

To overcome these limitations, we propose a novel approach that explicitly encourages exploration through differentiated temperature control during token sampling. Unlike previous strategies, our method maintains synchronous updates across all tokens, applying distinct temperature values tailored to token types: higher temperatures for reasoning tokens to promote exploration and lower temperatures for knowledge tokens to preserve accuracy. Additionally, we conduct an extensive analysis of token-level temperature sampling and multi-temperature sampling comprehensively. Experimental results across several challenging reasoning benchmarks demonstrate that our token-level sampling method substantially improves the reasoning performance of Qwen2.5-1.5B-Math by +6\% on AIME24, +1\% on AIME25, and +4.8\% on Minerva without extra computational cost.
\begin{figure*}[t]
    \centering
    \includegraphics[width=1\linewidth]{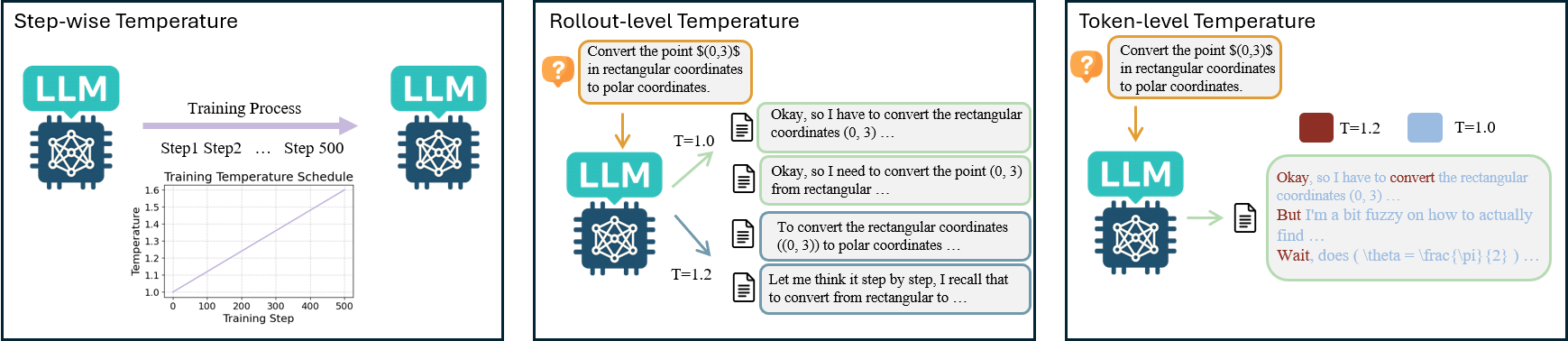}
    \caption{Different Multiple temperature control in Reinforcement learning process for LLMs.}
    \label{fig:main}
    \vspace{-0.2in}
\end{figure*}

\section{Related Works}

\subsection{Entropy Exploration in RL LLMs}
\vspace{-0.05in}
Recent work increasingly frames policy entropy as a first-class signal for exploration in RL-tuned LLMs. \citet{cui2025entropy} document how entropy collapses early in training, coinciding with plateaus, and propose interventions such as Clip-Cov and KL-Cov to preserve exploration dynamics. Building on this perspective, \citet{cheng2025reasoning} introduce an entropy-based advantage shaping method for PPO/GRPO, where a clipped, gradient-detached entropy term selectively amplifies updates on high-entropy “pivotal” tokens and reflective actions, leading to improved Pass@K on reasoning tasks. Beyond token-level shaping, \citet{wang2025beyond} demonstrate that a small set of high-entropy “forking” tokens disproportionately drives effective credit assignment, showing that updating only 20\% of tokens suffices, while low-entropy tokens hurt performance. Complementing this, \citet{wang2025reinforcement} show that RLVR can achieve strong improvements with a single training example, where entropy-aligned exploration yields extreme sample efficiency, e.g., entropy loss alone boosting MATH500 accuracy by over 27\%. In parallel, \citet{wang2025stabilizing} propose Dual-Token Constraints (Archer), which impose stronger constraints on low-entropy knowledge tokens and weaker ones on high-entropy reasoning tokens, thereby stabilizing factual knowledge while sustaining exploratory reasoning. Methodologically, \citet{deng2025decomposing} analyze the entropy–performance exchange in RLVR across stages and positions, introducing reward-adjustment strategies based on perplexity and token location rather than generic entropy regularization. Finally, \citet{wu2025invisible} in The Invisible Leash argue that RLVR is fundamentally constrained by the base model’s support—unable to explore beyond zero-probability outputs—and highlight an entropy–reward trade-off, recommending explicit exploration mechanisms to break this structural limitation. 

\subsection{Generation Diversity}
Generation diversity is important in text generation~\citep{chung2023increasing,du2025optimizing}, code generation~\citep{chon2024functional,lee2025diversely}, math solutions~\citep{xu2025not,yao2025diversity} in both \textbf{inference stage} and \textbf{training stage}. To improve the diversity in the \textbf{inference stage}, recent efforts have begun to move beyond static decoding hyperparameters, proposing adaptive sampling strategies that adjust at the token level. For instance, temperatures are dynamically selected per token or per sequence, balancing creativity with factual accuracy in open-ended generation tasks~\citep{dhuliawala2024adaptive,zhang2024edt}. Similarly, \citet{li2024dynamic} for dialogue adapts decoding choices according to conversational context, switching between exploratory and conservative modes when appropriate. In code generation, \citet{he2025towards} leverages token-level uncertainty to pause and rerank predictions, mitigating errors at high-entropy steps. However, recent work~\cite{verine2025improving} demonstrates that temperature scaling alone fails to improve diversity, highlighting the need to modify the training stage to effectively enhance diversity. Beyond inference-time adjustments, a growing line of work targets \textbf{training-time} mechanisms to elicit controllable diversity. CDE~\citep{dai2025cde}  treat the model’s own \emph{perplexity} as an actor-side curiosity bonus and pair it with a critic-side value-variance bonus, improving exploration efficiency under RLVR. DARLING~\citep{li2025jointly} explicitly optimizes a joint objective of quality \emph{and} semantic diversity via a learned partition/classifier over rollouts; notably, they show that naively rewarding lexical $n$-gram diversity can \emph{hurt} math pass@1 and be gamed by off-topic text, underscoring the need for semantic signals. For reasoning RL, \citet{song2025outcome} propose \emph{outcome-based exploration}, including a batch-level penalty on repeated outcomes within the same prompt to prevent diversity collapse while preserving accuracy. Complementarily, \citet{chen2025pass} incorporate \emph{Pass@k} directly into the reward and derive an advantage formulation that adaptively balances exploration and exploitation, boosting exploration without sacrificing solution quality. Together, these studies shift the focus from decoding heuristics to principled training-time interventions that make diversity both effective and verifiable.

\section{Methodology}
\label{sec:method}
In this work, we introduce an adaptive token-level temperature scheduling method that adjusts the sampling temperature based on the entropy of each token during generation. This helps the model explore more when uncertainty is high and focus more when confidence is high. Instead of using a single fixed temperature for all tokens, our approach dynamically changes it throughout the sequence. We also propose multi-temperature sampling per prompt, which samples outputs under several temperatures at once to reduce the risk of picking a bad temperature and approximate the best one without exhaustive tuning.

\subsection{Preliminary}

Group Relative Policy Optimization (GRPO)~\citep{shao2024deepseekmath} is introduced as an alternative to traditional value-based methods, specifically targeting the advantage estimation in policy gradient approaches. Unlike Proximal Policy Optimization (PPO)~\citep{schulman2017proximal}, which typically learns a separate value model, GRPO circumvents the need for value function approximation by generating multiple candidate responses per input prompt. Given a prompt $q$, GRPO samples a set of candidate responses $\{o^1, o^2, \dots, o^G\}$ and evaluates their respective rewards $\{R^1, R^2, \dots, R^G\}$. The advantage for each response is computed within this sampled set as:

{\footnotesize
\begin{equation}
    \hat{A}_t^i = \frac{R^i - \text{mean}(\{R^j\}_{j=1}^{G})}{\text{std}(\{R^j\}_{j=1}^{G})},
\end{equation}
where the mean and standard deviation are calculated within the sampled group. The loss for GRPO can thus be formulated as:
\begin{multline}
\mathcal{J}_{\text{GRPO}}(\theta) 
= \mathbb{E}_{q\sim \mathcal{D}, \{o^i\}_{i=1}^{G}\sim \pi_{\theta_{\text{old}}}(\cdot|q)} \Bigg[
\frac{1}{G}\sum_{i=1}^{G}\frac{1}{|o^i|}\sum_{t=1}^{|o^i|} \\
\min\! \Big(r_t^i(\theta)\hat{A}_t^i, \text{clip}\!\left(r_t^i(\theta), 1-\epsilon, 1+\epsilon\right)\hat{A}_t^i\Big) \\
 - \beta D_{\text{KL}}(\pi_{\theta}\|\|\pi_{\text{ref}})\Bigg].
\end{multline}
}

where the importance sampling ratio $r_t^i$ is defined by $r_t^i=\frac{\pi_{\theta}(o_t^i|q,o_{<t}^i)}{\pi_{\theta_{\text{old}}}(o_t^i|q,o_{<t}^i)}$, and $\beta$ is a hyperparameter weighting the KL-divergence regularization term between the current policy $\pi_\theta$ and a reference policy $\pi_{\text{ref}}$.

Decoupled Clip and Dynamic Sampling Policy Optimization (DAPO)~\citep{yu2025dapo} extends the GRPO framework by introducing a combination of novel techniques: dynamic sampling, token-level policy gradient optimization, clip-higher, and reward shaping strategies. Similar to GRPO, DAPO also generates multiple candidate responses per prompt, yet it further optimizes policy learning by dynamically adjusting clipping ranges and sampling strategies. Specifically, DAPO's optimization objective is defined as:
\begin{multline}
\mathcal{J}_{\text{DAPO}}(\theta) 
= \mathbb{E}_{(q,a)\sim\mathcal{D}, \{o^i\}_{i=1}^{G}\sim\pi_{\theta_{\text{old}}}(\cdot|q)} \\
\Bigg[
\frac{1}{G}\sum_{i=1}^{G}\frac{1}{|o^i|}\sum_{t=1}^{|o^i|} 
\min\!\Big(r_t^i(\theta)\hat{A}_t^i, \, 
\text{clip}\!\big(r_t^i(\theta), \\
1-\epsilon_{\text{low}},1+\epsilon_{\text{high}}\big)\hat{A}_t^i\Big)
\Bigg].
\end{multline}

subject to the constraint:
\begin{equation}
0 < |\{i \in \{1,\dots,G\} \mid \text{is\_equivalent}(o^i,a)\}| < G,
\end{equation}
where $\epsilon_{\text{low}}$ and $\epsilon_{\text{high}}$ denote the lower and upper bounds of the dynamic clipping range, respectively.

\subsection{Entropy-Based Token-level Temperature Scheduling}

Our method introduces a dynamic temperature mechanism guided by the entropy values of individual tokens during generation. Specifically, we calculate the entropy $H_t$ for each token $v$ generated at time step $t$:

\begin{equation}
H_t = -\sum_{v \in V} \pi_\theta(v | q, o_{<t}) \log \pi_\theta(v | q, o_{<t}),
\end{equation}

where $V$ denotes the vocabulary set, and $\pi_\theta(v | q, o_{<t})$ is the probability of token $v$ given the prompt $q$ and previously generated tokens $o_{<t}$ under the current policy $\pi_\theta$.

We maintain a fixed-size queue $Q$ of length $N$ to keep track of the entropy values for the most recent $N$ tokens. The threshold entropy $H_{\text{th}}$ is then defined dynamically as the entropy value at the $k\%$ quantile within this queue, sorted in descending order:

\begin{equation}
H_{\text{th}} = \text{Quantile}_{k\%}(Q).
\end{equation}

During token generation, we apply a two-tier temperature scheme based on the calculated entropy threshold:

\begin{equation}
\tau_t = \begin{cases}
T_{\text{high}}, & \text{if } H_t \geq H_{\text{th}}, \\
T_{\text{low}}, & \text{otherwise},
\end{cases}
\end{equation}

where $T_{\text{high}}$ is set to promote greater exploration for tokens with high uncertainty (higher entropy), and $T_{\text{low}}$ is chosen to ensure stable exploitation for tokens exhibiting lower uncertainty.

The entropy queue $Q$ is updated continuously during the generation process. Each newly computed token entropy $H_t$ is appended to the queue, and the oldest entropy value is removed once the queue exceeds its fixed length $N$. Formally, this update mechanism can be described as:

\begin{equation}
Q \leftarrow Q[2:N] \cup \{H_t\}, \quad \text{if } |Q| = N.
\end{equation}

The adaptive temperature scheduling described above is seamlessly integrated into the token sampling step, influencing the probability distribution used to generate each token:

\begin{equation}
\pi_{\theta, \tau_t}(o_t | q, o_{<t}) = \frac{\exp(\log \pi_\theta(o_t | q, o_{<t}) / \tau_t)}{\sum_{v \in V} \exp(\log \pi_\theta(v | q, o_{<t}) / \tau_t)}.
\end{equation}

By dynamically adjusting the sampling temperature based on real-time entropy metrics, our method effectively balances exploration and exploitation, adapting to the varying entropies of token predictions throughout the generation process.

\subsection{Multi-Temperature Sampling per Prompt}

Existing work~\citep{Polaris2025} shows that sampling temperature strongly influences model quality, yet practitioners~\citep{liu2025scaling} still choose $\tau$ by rule-of-thumb grid searches, an approach that is particularly costly in reinforcement-learning (RL) settings. Even schemes that predict a ``best" temperature after a few warm-up steps may be unreliable, because the relative ranking of temperatures can flip as training progresses and early-stage performance differences are often negligible.

To address this, beyond our entropy-driven scheduling policy, we introduce multi-temperature sampling: for every prompt the policy simultaneously generates candidate responses under several temperatures, allowing RL to pick the best answer from a richer, more diverse pool. Some temperatures may only shine at specific training stages; using many at once guarantees that high-quality samples remain available throughout. Moreover, the heterogeneity of sampling distributions enlarges the exploration space and improves data efficiency.

Formally, given a prompt $q$, instead of sampling responses $\{o^i\}_{i=1}^{n}$ with a single fixed temperature, we sample multiple subsets of responses, each with distinct temperatures $\{\tau_1, \tau_2, \dots, \tau_m\}$: $\{o^{(\tau_j)}\}_{j=1}^{m}, \quad$ where each subset  $o^{(\tau_j)} = \{o_i^{(\tau_j)}\}_{i=1}^{n_j}, 
\quad \sum_{j=1}^{m} n_j = n$.

This multi-temperature approach allows for exploration at multiple granularities, effectively capturing different aspects of the probability distribution characterized by the policy.

\vspace{-0.5em}
\section{Experiments}
\label{sec:experiments}

\subsection{Setup}
\textbf{Datasets.}
For the base model Qwen2.5-1.5B, we employ the MATH dataset~\citep{hendrycks2021measuring} comprising approximately 7,500 training instances, which is simple and effective to train a base model, selecting the standard MATH500 subset for testing performance metrics. To assess performance under more challenging conditions, we utilize the DAPO dataset~\citep{yu2025dapo}, containing 17K training instances specifically curated to encompass complex mathematical reasoning problems, and adopt AIME24 as the corresponding rigorous test benchmark.

\noindent\textbf{Baselines.}
Initially, the base model is trained using Group Relative Policy Optimization (GRPO) exclusively on the MATH dataset to establish foundational performance. Subsequently, we apply the DAPO algorithm on the DAPO dataset to further train the base model and observe performance dynamics. Finally, we implement Archer, the current state-of-the-art (SOTA) method which integrates entropy constraints into the training process, to further refine the model's capability and achieve state-of-the-art performance at the 1.5B parameter scale.

\noindent\textbf{Implementation Details.}
Our reinforcement learning (RL) experiments are conducted within the VeRL framework. The validation is based on Math-Verify\footnote{https://github.com/huggingface/Math-Verify}. We set Temperature to 0.2 at avg@256 to fully activate the ability of model in base model settings and 0.6 for pass@256 to explore the diversity of solution. For the DAPO-based methods, we define clipping thresholds specifically as $\epsilon_{\text{low}}=0.2$ and $\epsilon_{\text{high}}=0.28$, excluding KL penalty and entropy regularization terms from the loss formulation to simplify the optimization objective. Each prompt undergoes 8 rollout samplings for our methods and baselines, maintaining a maximum response length of 2048 tokens for the base model and extending to 8192 tokens for the DeepSeek-R1-Distill-Qwen-1.5B model to accommodate increased complexity. Training is executed on high-performance GPU clusters; the base models leverage a single computational node, while the DeepSeek variant is trained using two nodes, each equipped with 8 NVIDIA H100 GPUs with 80GB memory capacity, ensuring sufficient computational resources for extensive training. For Archer, we adopt the default settings in their paper.

\subsection{Token Level Temperature}
\begin{table*}[t]
\caption{Evaluation results of Token-level temperature sampling and single temperature on math reasoning benchmarks. }
\label{tab:math_trimmed}
\centering
\setlength{\tabcolsep}{5pt}
\begin{tabular}{l *{7}{c}} 
\toprule
\multirow{2}{*}{\textbf{Method}} &
\multicolumn{2}{c}{\textbf{AIME24}} &
\multicolumn{2}{c}{\textbf{AIME25}} &
\multicolumn{1}{c}{\textbf{Minerva}} &
\multicolumn{1}{c}{\textbf{Olympiad}} &
\multirow{2}{*}{\textbf{Avg.}}\\
\cmidrule(lr){2-3}\cmidrule(lr){4-5}\cmidrule(lr){6-6}\cmidrule(lr){7-7}
 & avg@256 & pass@256 & avg@256 & pass@256 & pass@1 & pass@1 & \\
\midrule
Base Model      & 0.093   & 0.560    & 0.058   & 0.444    & 0.176   & 0.322   &  0.162  \\
DAPO-T1.0       & 0.180 & 0.548 & 0.099 & 0.517 & 0.287 & 0.360 & 0.231 \\
DAPO-T1.2       & 0.175 & \textbf{0.574} & 0.106 & \textbf{0.533} & 0.319 & \textbf{0.382} & 0.246 \\

Ours            & \textbf{0.235} & 0.554 & \textbf{0.115} & 0.405 & \textbf{0.335} & \textbf{0.382} & \textbf{0.267} \\
\bottomrule
\end{tabular}
\vspace{-0.1in}
\end{table*}

Table \ref{tab:math_trimmed} shows that our token-level temperature strategy delivers clear gains over both temperature-tuned DAPO baselines. Relative to DAPO-T1.0 (DAPO with temperature = 1.0), our model lifts the AIME24 average from 0.180 to 0.235 (+30 \%), AIME25 from 0.099 to 0.115 (+17 \%), Minerva from 0.287 to 0.335 (+17 \%) and Olympiad from 0.360 to 0.382 (+6 \%), giving an overall improvement of 0.036 absolute points (around 16 \%) across the four benchmarks. The similar results can be observed compared with DAPO-T1.2 (DAPO with temperature = 1.2). While our Token-level temperature sampling boosts the average accuracy across all four benchmarks, it does introduce a trade-off in the high-temperature pass@256 metric. In particular, on AIME25 the pass@256 rate falls from 0.517 with DAPO-T1.0 and 0.533 with DAPO-T1.2 to 0.405 with our method. We attribute this drop to the stronger exploration encouraged by the token-level temperature: the model now spreads probability mass over a wider range of solution paths, which raises the chance that its top-ranked answer (pass@1) is correct, yet at 256-sample temperature some questions lose previously covered but lower-probability solution variants, reducing the cumulative pass rate. These consistent improvements on pass@1 demonstrate the effectiveness of our token-level temperature sampling which improves the performance of single temperature.

\subsection{Multi-temperature per questions}
\begin{figure*}[t]
    \centering
    \begin{minipage}[t]{0.56\textwidth}
        \centering
        \includegraphics[width=\textwidth]{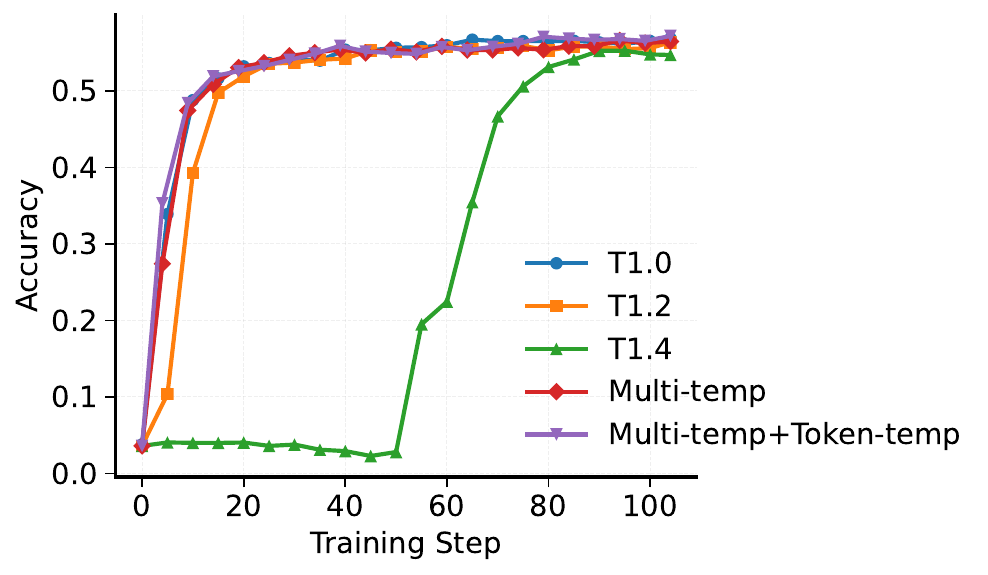}
    \end{minipage}
    \hfill
    \begin{minipage}[t]{0.43\textwidth}
        \centering
        \includegraphics[width=\textwidth]{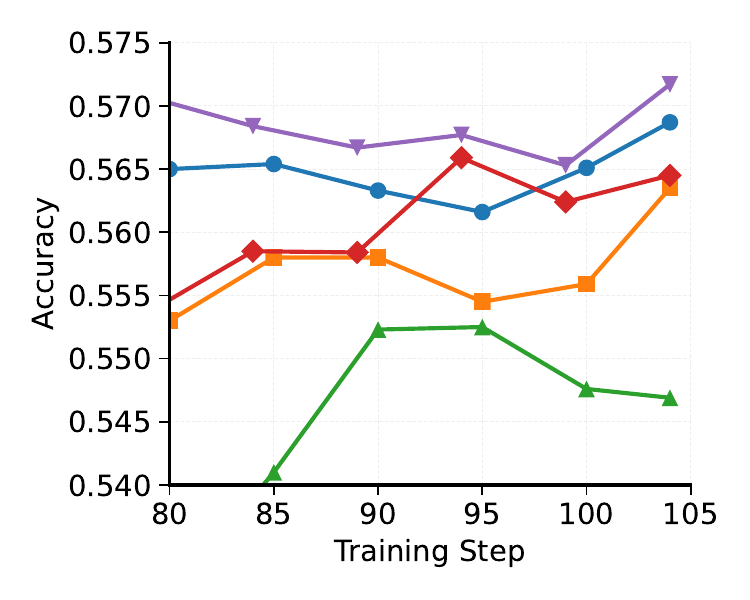}
    \end{minipage}
    \caption{Model performance of Multiple temperature and the combination of Multiple temperature and Token-level temperature.}
    \label{fig:MATH_it}
    \vspace{-0.2in}
\end{figure*}
\vspace{-0.1in}
\noindent The DAPO dataset is difficult for the vanilla \emph{base} model Qwen2.5-1.5B; consequently, we resort to the \textsc{MATH} dataset, which remains non-trivial yet permits stable optimisation. A dense sweep over single–temperature settings shows that satisfactory convergence occurs only for $\tau<1.4$, while hotter values (e.g., $\tau>1.4$) reach competitive performance (e.g., $Acc>0.1$) omitted in the figure. Guided by this bound, we examine two multi-temperature regimes: (i) an \emph{in-range mixture}, where each sampler draws from ${1.0,1.1,1.2,1.3,1.4}$, and (ii) a \emph{mixed-range mixture}, in which one sampler remains within range ($\tau=1.0$) and the others are intentionally set above the threshold ($\tau>1.4$). As illustrated in Fig.~\ref{fig:MATH_it}, the in-range mixture tracks the fastest single-temperature baseline early on and surpasses it at step 95. Introducing the combination of our \emph{token-level temperature sampling} and \emph{Multiple temperature sampling} on top of this mixture yields a further rise to $0.571$ and attenuates late-stage variance. These findings indicate that a single well-calibrated sampler can anchor learning even amid overly explorative counterparts, and that fine-grained, token-level temperature modulation can harness additional exploration without sacrificing—indeed, while modestly improving—ultimate performance.

\begin{figure}[t]
    \centering
    \begin{minipage}[t]{0.35\textwidth}
        \centering
        \includegraphics[width=\textwidth]{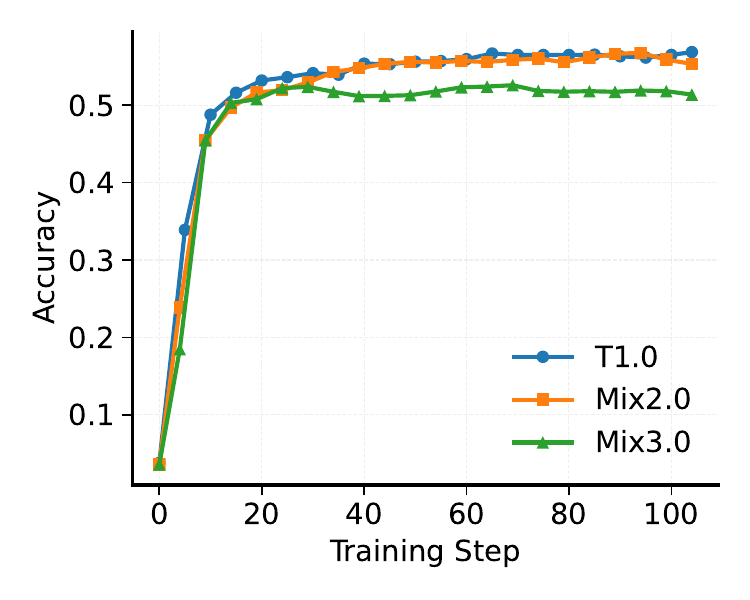}
    \end{minipage}
    \hfill
    \begin{minipage}[t]{0.35\textwidth}
        \centering
        \includegraphics[width=\textwidth]{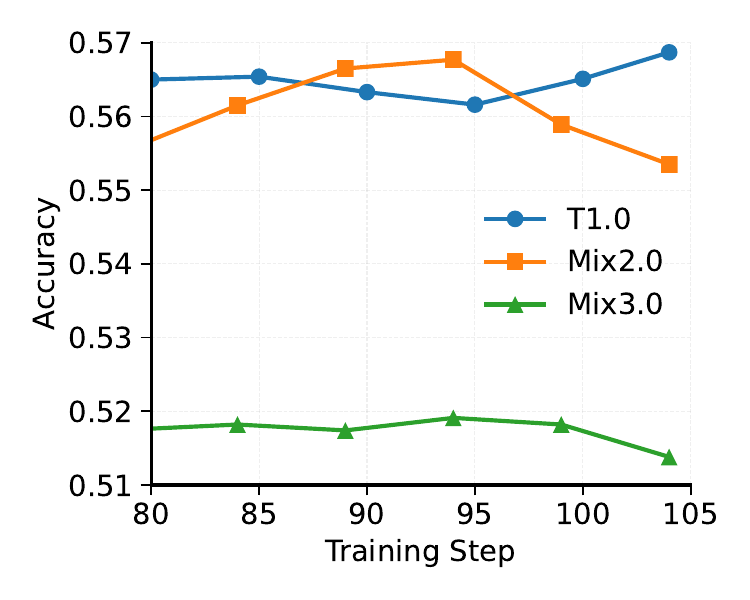}
    \end{minipage}
    \caption{Model performance of Multiple temperatures on out-of-range temperatures.}
    \label{fig:oot}
\end{figure}

To evaluate the resilience of our multi-temperature sampling scheme under extreme conditions, we combined the baseline setting $\tau = 1.0$ with two sets of markedly higher temperatures: $\{1.8,1.9,2.0,2.1\}$ and $\{2.8,2.9,3.0,3.1\}$. Despite having four of the five temperatures outside the empirically stable range, the first group ($\tau\in[1.8, 2.1]$) matches the performance of the single-temperature best at $\tau = 1.0$, which is shown in Fig.~\ref{fig:oot}. Even the second, more aggressive group ($\tau\in[2.8, 3.1]$) secures a score of 0.51—only a modest decline from the optimal 0.56—demonstrating that multi-temperature sampling remains effective under extremely high-temperature regimes.

\subsection{Combine with other strategy.}

\begin{figure}[t]
    \centering
    \begin{subfigure}[t]{0.35\textwidth}
        \centering
        \includegraphics[width=\textwidth]{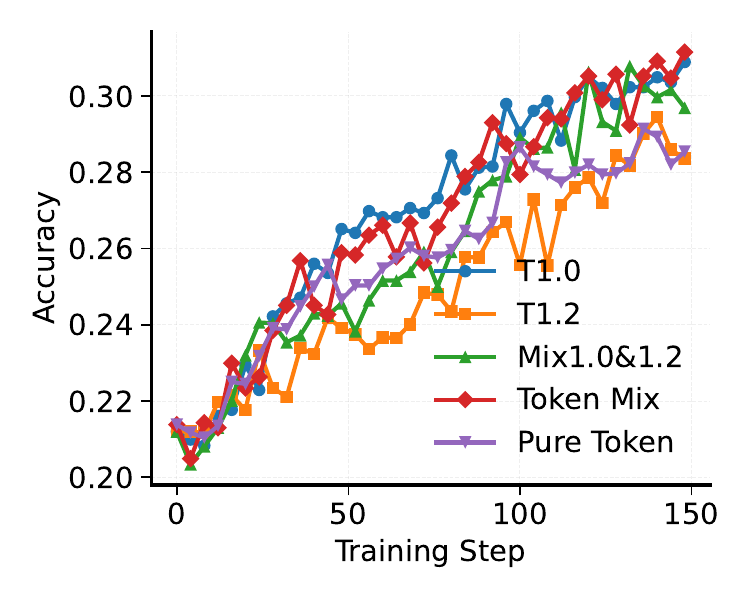}
        \caption{Pass@1}
        \label{fig:image1}
    \end{subfigure}
    \hfill
    \begin{subfigure}[t]{0.35\textwidth}
        \centering
        \includegraphics[width=\textwidth]{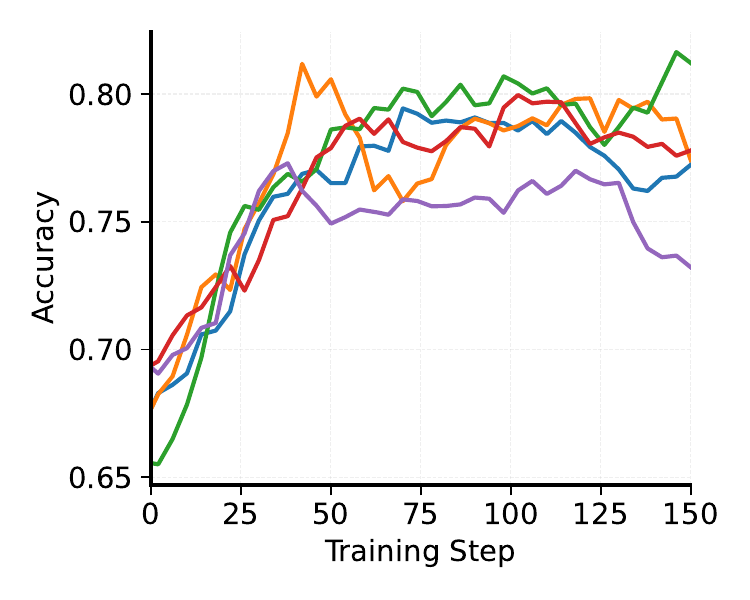}
        \caption{Pass@128}
        \label{fig:image2}
    \end{subfigure}
    \caption{Performance of Deepseek-1.5B trained on Baseline temperature and proposed methods. We smooth the Pass@128 for visualization.}
    \label{fig:deepseek1.5b}
    \vspace{-0.2in}
\end{figure}

Archer~\citep{wang2025stabilizing} is a state-of-the-art (SOTA) method that incorporates entropy-based constraints, selectively applying different treatments to high-entropy and low-entropy tokens. In this work, we investigate whether incorporating token-level temperature control in conjunction with these constraints can further enhance exploration during training. Our results show that Archer achieves the best Pass@1 performance when using a fixed temperature of $1.0$, while the highest Pass@128 is observed at a temperature of $1.2$. Based on these observations, we set the token-level temperature with $T_{\text{high}} = 1.2$, $T_{\text{low}} = 1.0$, and a high-entropy token ratio of $k=40\%$. For multiple temperature sampling, we use the set $\{1.0, 1.2\}$.

The results indicate that multiple temperature sampling achieves a Pass@1 performance that lies between the values achieved by fixed temperatures of $1.0$ and $1.2$, and is closer to the better-performing temperature $1.0$ after step 100. In terms of Pass@128, temperature mixing surpasses the fixed setting of $T=1.2$. Interestingly, we observe that token-level temperature achieves competitive performance in early training (before step 100), but experiences a sharp decline after, despite training rewards remaining consistent with the $T=1.0$ baseline. Upon further inspection, we find that the model often repeats a specific reasoning step or logical phrase without progressing, potentially due to a mismatch between the temperature used during training and evaluation.

To address this, we apply temperature mixing at the token level, assigning half of the tokens $T_{\text{high}} = 1.2$ and the other half $T_{\text{low}} = 1.0$. This combined strategy mitigates the degeneration issue and recovers strong performance: Pass@1 matches the $T=1.0$ baseline, while Pass@128 surpasses it. These findings suggest that both token-level temperature sampling and multiple temperature sampling contribute to better exploration by leveraging higher temperatures, while maintaining stability through lower-temperature sampling.

\vspace{-0.1in}
\subsection{On-policy vs Off-policy}
\begin{figure}[t]
    \centering
    \begin{subfigure}[b]{0.23\textwidth}
        \includegraphics[width=\linewidth]{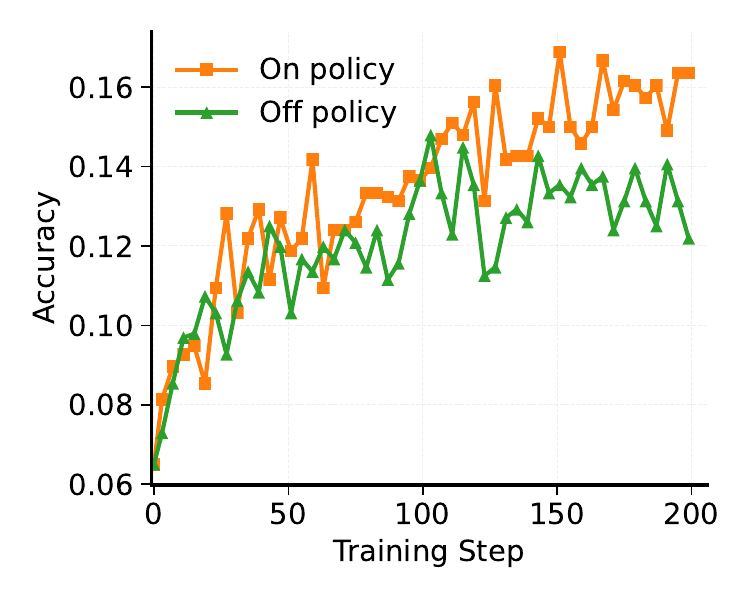}
        \caption{Performance}
        \label{fig:sub1}
    \end{subfigure}
    \hfill
    \begin{subfigure}[b]{0.23\textwidth}
        \includegraphics[width=\linewidth]{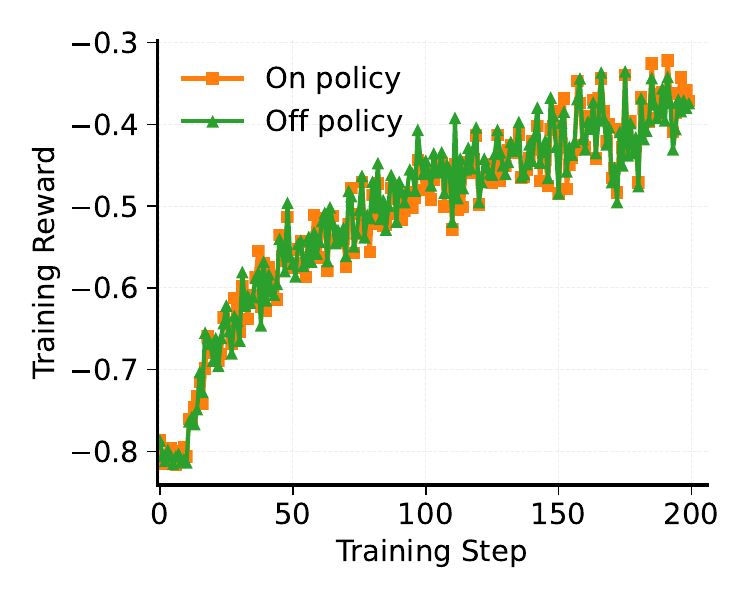}
        \caption{Training Reward}
        \label{fig:sub2}
    \end{subfigure}
    \hfill
    \begin{subfigure}[b]{0.23\textwidth}
        \includegraphics[width=\linewidth]{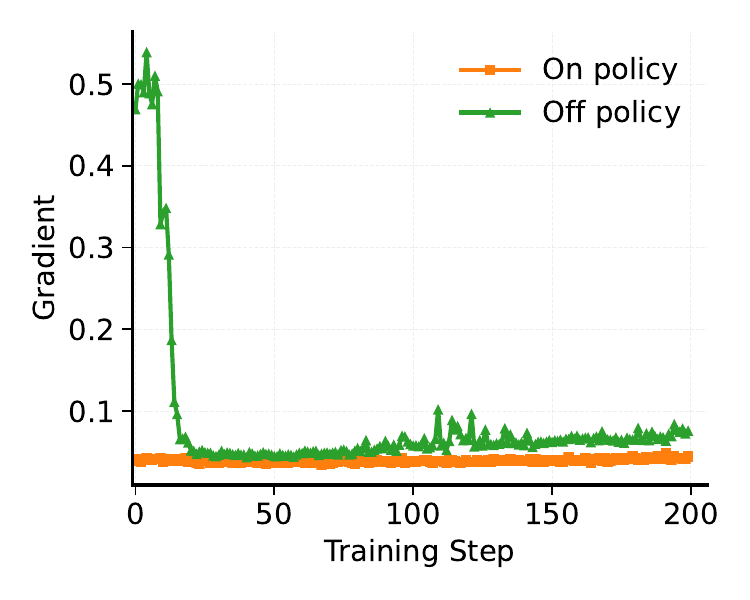}
        \caption{Gradient Norm}
        \label{fig:sub3}
    \end{subfigure}
    \hfill
    \begin{subfigure}[b]{0.23\textwidth}
        \includegraphics[width=\linewidth]{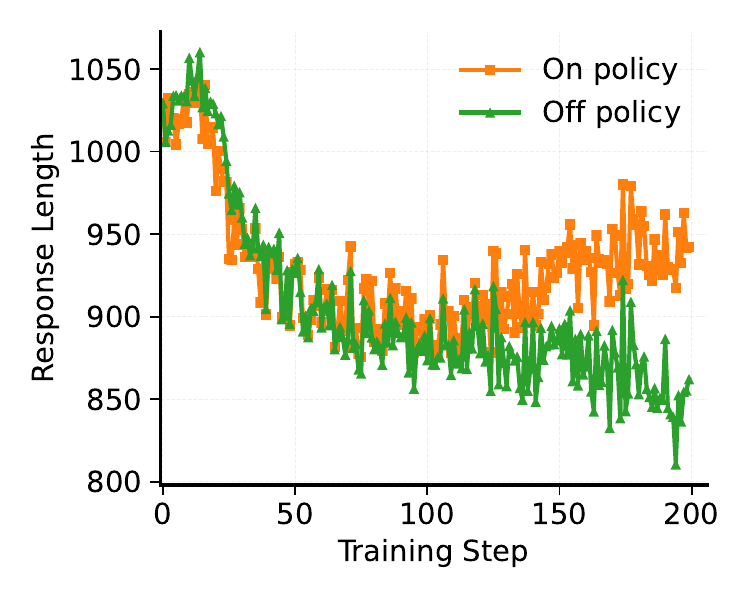}
        \caption{Response Length}
        \label{fig:sub4}
    \end{subfigure}
    \caption{On-policy vs Off-policy on Token-level temperature training.}
    \hfill
    \label{fig:policy}
    \vspace{-0.3in}
\end{figure}

\vspace{-0.1in}
To investigate the role of on-policy updates during training, we perform both on-policy and off-policy training at the token-level temperature sampling. Specifically, we apply token-wise temperature scaling during the policy training phase to ensure on-policy behavior, while using a fixed base temperature during the policy update phase to simulate off-policy training. As shown in the figure~\ref{fig:policy}, the on-policy strategy yields better performance and a more stable training process, particularly in the early stages when the gradient norm is high. The on-policy approach results in improved answer quality and longer response lengths.

\subsection{Step-wise multiple temperature}
\vspace{-0.1in}
Prior work suggests that progressively increasing the sampling temperature during training can boost model performance, yet the optimal scheduling strategy remains unclear. We investigate two approaches for raising the temperature: (i) a spike schedule that increments the temperature by a fixed amount at regular intervals, and (ii) a linear schedule that grows the temperature continuously. When the temperature of all tokens is raised, a spike of +0.1 every 50 steps performs comparably to the linear schedule on both Pass@1 and Pass@32. In contrast, a finer-grained spike of +0.05 every 25 steps degrades both accuracy and output length. Entropy traces reveal that the latter schedule injects additional uncertainty before the model has fully exploited the exploration induced by the previous spike, preventing the system from reaching equilibrium.

For token-level temperature scaling, a spike of +0.1 every 50 steps outperforms the alternatives, yielding the longest average response length (1 025 tokens) and the highest Pass@1, while the linear schedule attains the best Pass@32 and remains competitive on Pass@1. Taken together, these findings indicate that the model tolerates larger, less frequent temperature jumps once it has entered a stable low-entropy regime. When computational resources are limited and extensive hyper-parameter sweeps are impractical, a linear schedule remains a conservative and robust choice.

\begin{figure}[t]
    \centering
    \begin{subfigure}[t]{0.23\textwidth}
        \centering
        \includegraphics[width=\textwidth]{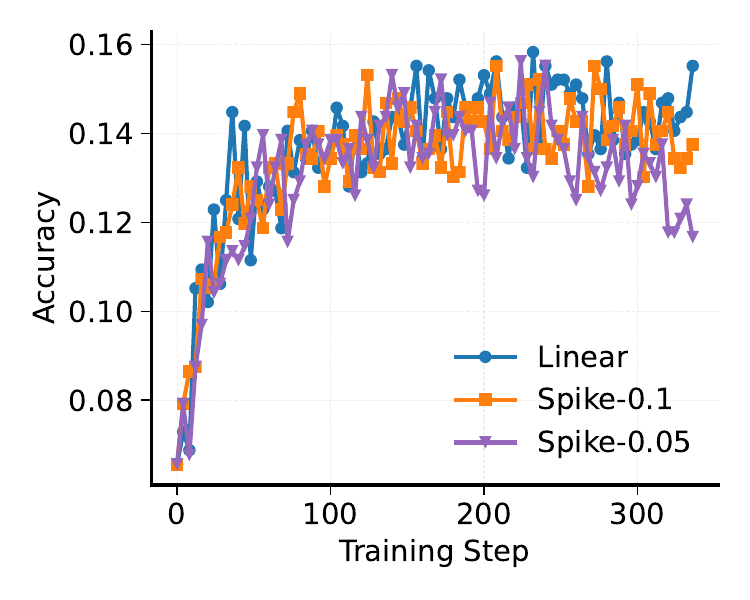}
        \caption{Pass@1}
        \label{fig:image1}
    \end{subfigure}
    \hfill
    \begin{subfigure}[t]{0.23\textwidth}
        \centering
        \includegraphics[width=\textwidth]{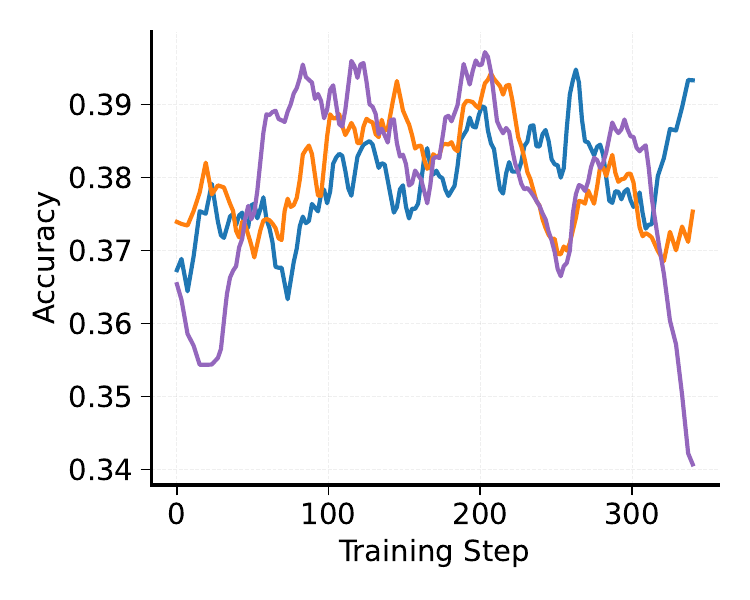}
        \caption{Pass@32}
        \label{fig:image2}
    \end{subfigure}
    \hfill
    \begin{subfigure}[t]{0.23\textwidth}
        \centering
        \includegraphics[width=\textwidth]{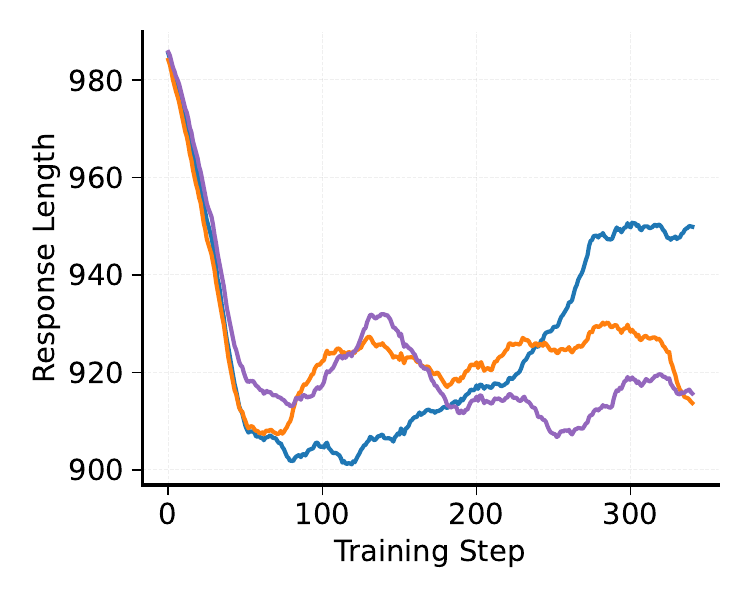}
        \caption{Response Length}
        \label{fig:length}
    \end{subfigure}
    \hfill
    \begin{subfigure}[t]{0.23\textwidth}
        \centering
        \includegraphics[width=\textwidth]{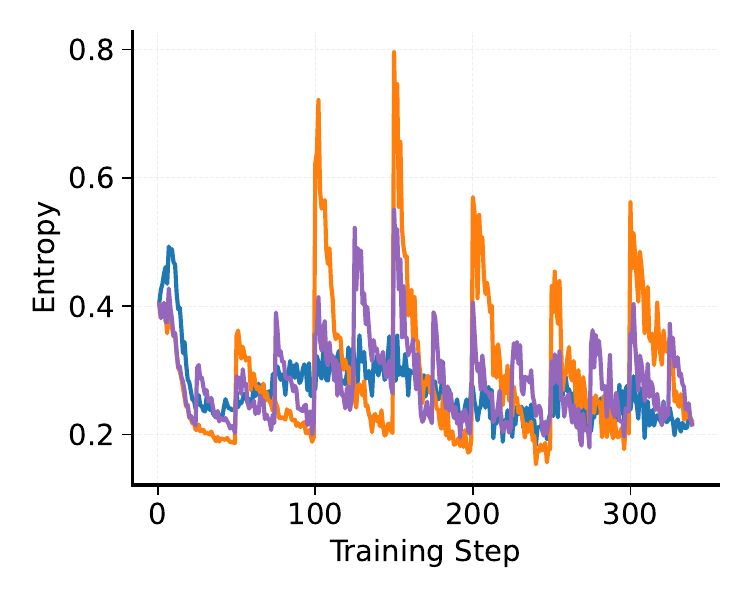}
        \caption{Entropy}
        \label{fig:entropy}
    \end{subfigure}
    \caption{Performance of increasing temperature with the process of training.}
    \label{fig:step_temp}
\end{figure}

\begin{figure}[t]
    \centering
    \begin{subfigure}[t]{0.23\textwidth}
        \centering
        \includegraphics[width=\textwidth]{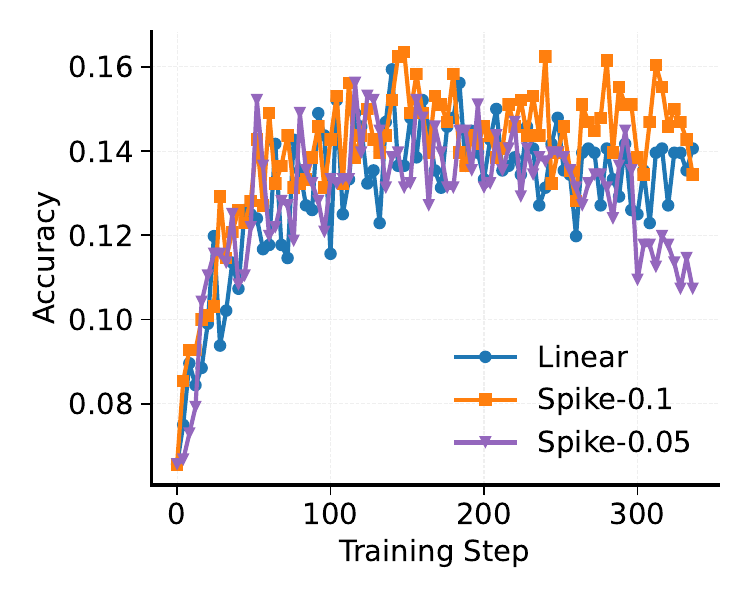}
        \caption{Pass@1}
        \label{fig:image1}
    \end{subfigure}
    \hfill
    \begin{subfigure}[t]{0.23\textwidth}
        \centering
        \includegraphics[width=\textwidth]{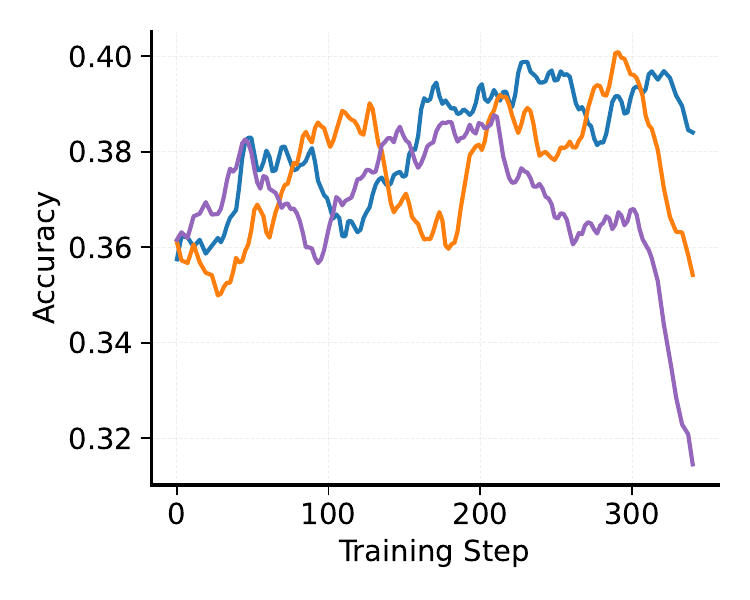}
        \caption{Pass@32}
        \label{fig:image2}
    \end{subfigure}
    \hfill
    \begin{subfigure}[t]{0.23\textwidth}
        \centering
        \includegraphics[width=\textwidth]{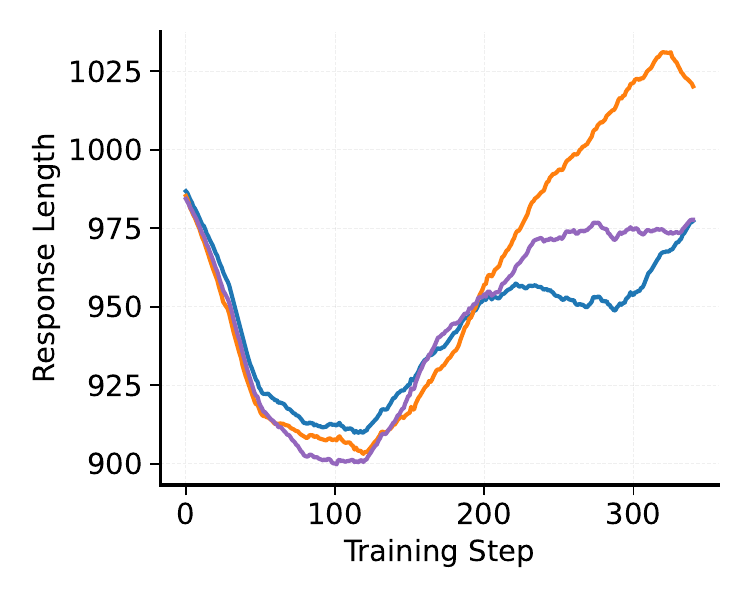}
        \caption{Response Length}
        \label{fig:length}
    \end{subfigure}
    \hfill
    \begin{subfigure}[t]{0.23\textwidth}
        \centering
        \includegraphics[width=\textwidth]{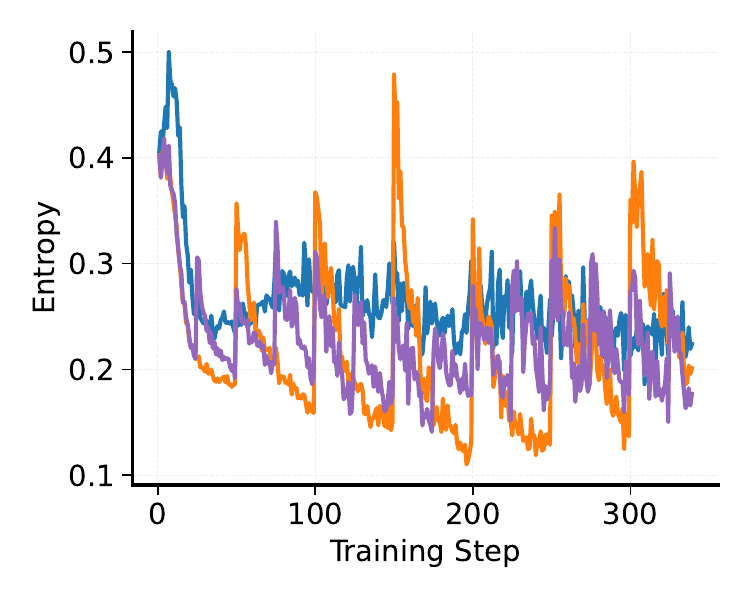}
        \caption{Entropy}
        \label{fig:entropy}
    \end{subfigure}
    \caption{Performance of increasing temperature with the process of token-level temperature training.}
    \vspace{-0.1in}
    \label{fig:step_temp_tokenlevel}
\end{figure}

\section{Conclusion}
In this work, we identified that existing research on multi-temperature strategies in RL-based LLM training remains limited. To address this gap, we proposed and systematically investigated both sample-level and token-level multi-temperature approaches. Our results show that sample-level multi-temperature can approximate the performance of the optimal temperature without prior knowledge of its value, while token-level multi-temperature can even surpass the optimal temperature under short-text generation scenarios. Combining these two methods leads to more temperature-resilient training and consistently outperforms the baseline. Furthermore, we explored step-wise multi-temperature scheduling and found that well-timed spikes can yield notable improvements in both performance and response length, while linear increase of temperature can be a conservative strategy. We hope our findings offer new insights into effective temperature configuration for RL-based LLM training.

\section{Limitation}
Our experiments are limited to the 1.5B-parameter Qwen2.5 model due to computational resource constraints. We did not conduct evaluations on larger models or other architectures. As a result, our conclusions about the effectiveness of multi-temperature reinforcement learning may not fully generalize beyond this model size or family.

\section{Ethical Statement}
This work does not introduce new datasets or collect human subjects data. All benchmarks used—MATH, DAPO, AIME, and related datasets—are publicly available and contain only problem–solution pairs without personally identifiable information.

Our research aims to improve the reasoning capability and controllability of LLMs through enhanced exploration mechanisms in reinforcement learning.

\bibliography{iclr2025_conference}

\end{document}

%% file: math_commands.tex
\usepackage{amsmath,amsfonts,bm}

\def\eqref#1{equation~\ref{#1}}

\def\1{\bm{1}}

\DeclareMathAlphabet{\mathsfit}{\encodingdefault}{\sfdefault}{m}{sl}
\SetMathAlphabet{\mathsfit}{bold}{\encodingdefault}{\sfdefault}{bx}{n}

